\documentclass[]{article}
\usepackage[letterpaper]{geometry}
\usepackage{mtsummit2015}
\usepackage{times}
\usepackage{url}
\usepackage{latexsym}
\usepackage{natbib}
\usepackage{layout}
\usepackage[pdftex]{graphicx}
\usepackage{subcaption}
\usepackage{listings}

\begin{document}

\title{\bf Why not be Versatile? \\
  Applications of the SGNMT Decoder for Machine Translation}  

\author{\name{\bf Felix Stahlberg}$^\dag$ \hfill  \addr{fs439@cam.ac.uk}\\ 
		\name{\bf Danielle Saunders}$^\dag$ \hfill \addr{ds636@cam.ac.uk}\\ 
        \name{\bf Gonzalo Iglesias}$^{\ddagger}$ \hfill \addr{giglesias@sdl.com}\\ 
       \name{\bf Bill Byrne}$^{\ddagger\dag}$ \hfill \addr{bill.byrne@eng.cam.ac.uk, bbyrne@sdl.com}\\
        $^\dag$\addr{Department of Engineering, University of Cambridge, UK}\\
        $^{\ddagger}$\addr{SDL Research, Cambridge, UK}
}

\maketitle
\pagestyle{empty}

\begin{abstract}
  SGNMT is a decoding platform for machine translation which allows paring various modern neural models of translation with different kinds of constraints and symbolic models. In this paper, we describe three use cases in which SGNMT is currently playing an active role: (1) {\em teaching} as SGNMT is being used for course work and student theses in the MPhil in Machine Learning, Speech and Language Technology at the University of Cambridge, (2) {\em research} as most of the research work of the Cambridge MT group is based on SGNMT, and (3) {\em technology transfer} as we show how SGNMT is helping to transfer research findings from the laboratory to the industry, eg.\ into a product of SDL plc.
\end{abstract}

\section{Introduction}

\begin{figure}[!t]
\centering
\includegraphics[width=1\linewidth]{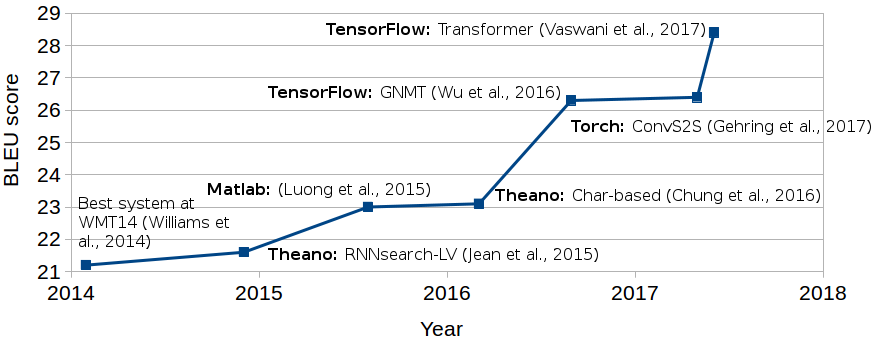}
\caption{Best systems on the English-German WMT {\em news-test2014} test set over the years (BLEU script: Moses' \texttt{multi-bleu.pl}).}
\label{fig:sota}
\end{figure}

The rate of innovation in machine translation (MT) has gathered impressive momentum over the recent years. The discovery and maturation of the neural machine translation (NMT) paradigm~\citep{sutskever,bahdanau} has led to steady and substantial improvements of translation performance~\citep{uedin-wmt14,rnnsearch-lv,luong-attention,char-nmt,gnmt,convs2s,transformer}. Fig.~\ref{fig:sota} shows that this progress is often driven by significant changes in the network architecture. This volatility poses major challenges in MT-related research, teaching, and industry. Researchers potentially spend a lot of time implementing to keep their setups up-to-date with the latest models, teaching needs to identify suitable material in a changing environment, and the industry faces demanding speed requirements on its deployment processes. Another practical challenge many researchers are struggling with is the large number of available NMT tools~\citep{blocks,marian,opennmt,nematus,neuralmonkey,mmt,sockeye}.\footnote{See \url{https://github.com/jonsafari/nmt-list} for a complete list of NMT software.} Committing to one particular NMT tool bears the risk of being outdated soon, as keeping up with the pace of research is especially costly for NMT software developers.

The open-source SGNMT (Syntactically Guided Neural Machine Translation) decoder\footnote{Full documentation available at \url{http://ucam-smt.github.io/sgnmt/html/}.}~\citep{sgnmt2} is our attempt to mediate the effects of the rapid progress in MT and the diversity of available NMT software. SGNMT introduces the concept of {\em predictors} as abstract scoring modules with left-to-right semantics. We can think of a {\em predictor} as an interface to a particular neural model or NMT tool. However, the interface also allows to implement constraints like in lattice or $n$-best list rescoring, and symbolic models such as $n$-gram language models or counting models as predictors. Our software architecture is designed to facilitate the implementation of new predictors. Therefore, SGNMT can be extended to a new model or tool with very limited coding effort because rather than reimplementing models it is often enough to access APIs within an adapter predictor.\footnote{Making all models of the T2T library~\citep{t2t} available to SGNMT took less than 200 lines of code.} Software packages which are not written in Python can be exposed in SGNMT if they have a Python interface.\footnote{For example, the neural language modeling software NPLM~\citep{nplm} is written in C++, but can be accessed in SGNMT via its Python interface.} Once a new predictor is implemented, it can be directly combined with all other predictors which are already available in SGNMT. Therefore, general techniques like lattice and $n$-best list rescoring~\citep{sgnmt1,nbest}, ensembling, MBR-based NMT~\citep{mbr}, etc.\ only need to be implemented once (as predictor), and are automatically available for all models. This does not only speed up the transition to a new NMT toolkit, it also allows the combination of different NMT implementations, eg.\ ensembling a Theano-based NMT model~\citep{blocks} with a TensorFlow-based Tensor2Tensor~\citep{t2t} model. \citet{eva-wordordering} demonstrated the versatility of SGNMT by combining five very different models (RNN LM, feedforward NPLM, Kneser-Ney LM, bag-to-seq model, seq-to-seq model) and a bag-of-words constraint using predictors. 

Not only the way scores are assigned to translations is open for extension in SGNMT (via predictors), but also the search strategy ({\em decoder}) itself. Decoders in SGNMT are defined upon the predictor abstraction, which means that any search strategy is compatible with any predictor constellation. Therefore, common search procedures like beam search do not need to be reimplemented for every new model or toolkit.

Secs.~\ref{sec:architecture} to~\ref{sec:output-formats} describe central concepts in SGNMT like predictors and decoders briefly and outline some common use cases.  Sec.~\ref{sec:research} shows that the SGNMT software architecture has proven to be very well suited for our research as new directions can be quickly prototyped, and new NMT toolkits can be introduced without breaking old code. Sec.~\ref{sec:teaching} and Sec.~\ref{sec:industry} discuss the benefits of SGNMT in teaching and industry, respectively.

\section{The Predictor Interface}
\label{sec:architecture}

Predictors in SGNMT provide a uniform interface for models and constraints. Since predictors are decoupled from each other, any predictor can be combined with any other predictor in a linear model. One predictor usually has a single responsibility as it represents a single model or type of constraint. Predictors need to implement the following methods:

\begin{itemize}
\item \texttt{initialize(src\_sentence)} Initialize the predictor state using the source sentence.
\item \texttt{get\_state()} Get the internal predictor state.
\item \texttt{set\_state(state)} Set the internal predictor state.
\item \texttt{predict\_next()} Given the internal predictor state, produce the posterior over target tokens for the next position.
\item \texttt{consume(token)} Update the internal predictor state by adding \texttt{token} to the current history.
\end{itemize}

\begin{figure}[!t]
\centering
\includegraphics[width=1\linewidth]{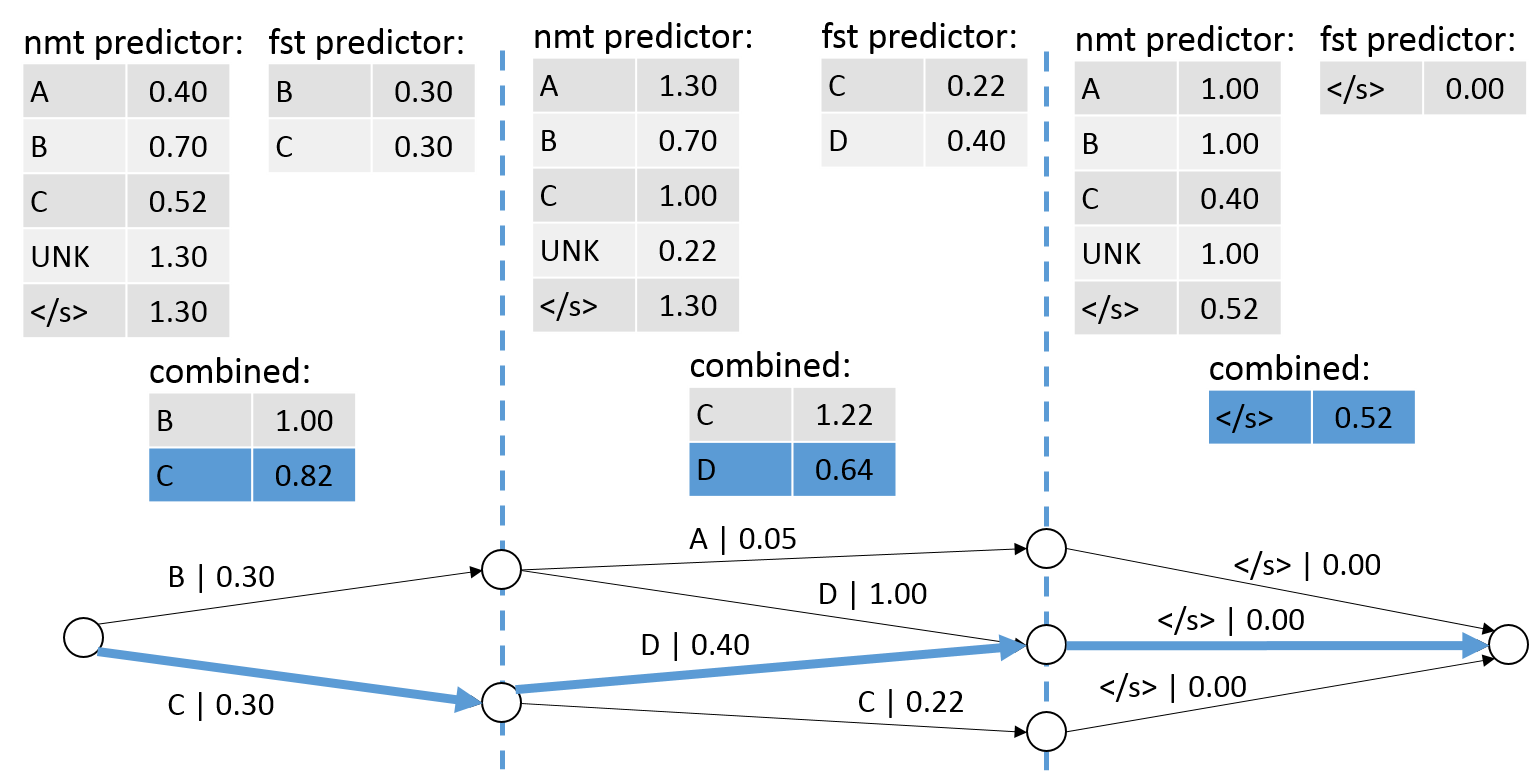}
\caption{Greedy decoding with the predictor constellation {\em nmt,fst} for lattice rescoring.}
\label{fig:nmt_fst}
\end{figure}

The structure of the predictor state and the implementations of these methods differ substantially between predictors. \citet{sgnmt2} provide a full list of available predictors. Fig.~\ref{fig:nmt_fst} illustrates how the {\em fst} and the {\em nmt} predictors work together to carry out (greedy) lattice rescoring with an NMT model. The \texttt{predict\_next()} method of the {\em nmt} predictor produces a distribution over the complete NMT vocabulary $\{\textrm{A}, \textrm{B}, \textrm{C}, \textrm{UNK}, \textrm{\textless/s\textgreater}\}$ at each time step in form of negative log probabilities. The {\em fst} predictor returns the scores of symbols with an outgoing arc from the current node in the FST in \texttt{predict\_next()}. The linear combination of both scores is used to select the next word, which is then fed back to the predictors via \texttt{consume()}. Words outside a predictor vocabulary are automatically matched with the UNK score. For instance, `D' in Fig.~\ref{fig:nmt_fst} is matched with the NMT `UNK' token. Pseudo-code for the predictors and the decoder is listed in Figs.~\ref{fig:predictor_code} and~\ref{fig:decoder_code}, respectively.

\begin{figure}[!t]
    \centering
    \begin{subfigure}[t]{0.49\textwidth}
        \footnotesize
        \begin{lstlisting}[language=Python]
class NMTPredictor(Predictor):
  def initialize(src_sentence):
    enc_states = enc_computation_graph(
        src_sentence)
    dec_input = [BOS]
  def predict_next():
    scores, dec_state = \
        dec_computation_graph(
            dec_input, enc_states)
    return scores
  def consume(word):
    dec_input = word
  def get_state():
    return dec_state, dec_input
  def set_state(state):
    dec_state, dec_input = state
\end{lstlisting}
        \caption{The {\em nmt} predictor}
        \label{fig:nmt_predictor}
    \end{subfigure}
    ~ 
    \begin{subfigure}[t]{0.49\textwidth}
        \footnotesize
\begin{lstlisting}[language=Python]
class FSTPredictor(Predictor):
  def initialize(src_sentence):
    Load FST file
    cur_node = start_node
          
  def predict_next():
    return outgoing_arcs(cur_node)
          
  def consume(word):
    cur_node = cur_node.arcs[word]
          
  def get_state():
    return cur_node
          
  def set_state(state):
    cur_node = state          
\end{lstlisting}
        \caption{The {\em fst} predictor}
        \label{fig:fst_predictor}
    \end{subfigure}
    \caption{Pseudo-code predictor implementations}\label{fig:predictor_code}
\end{figure}

\begin{figure}[!t]
\centering
\footnotesize
\begin{lstlisting}[language=Python]
          class GreedyDecoder(Decoder):
            def decode(src_sentence):
              initialize_predictors(src_sentence)
              trgt_sentence = []
              trgt_word = None
              while trgt_word != EOS:
                trgt_word = argmin(combine(predictors.predict_next()))
                trgt_sentence.append(trgt_word)
                predictors.consume(trgt_word)
              return trgt_sentence
\end{lstlisting}
\caption{Pseudo-code implementation of greedy decoding}
\label{fig:decoder_code}
\end{figure}

\section{Search Strategies}

Search strategies, called {\em Decoders} in SGNMT, search over the space spanned by the predictors. We use different decoders for different predictor constellations, e.g.\ heuristic search for bag-of-words problems~\citep{eva-wordordering}, or beam search for NMT. SGNMT can also be used to analyze search errors. Tab.~\ref{tab:decoders} compares five different search configurations for SMT lattice rescoring with a Transformer model~\citep{transformer} on a subset\footnote{SMT lattices are lightly pruned by removing paths whose weight is more than five times the weight of the shortest path. For the experiments in Tab.~\ref{tab:decoders} we removed very long sentences from the original test set to keep the runtime under control. Lattices have 271 nodes and 408 arcs on average.} of the Japanese-English Kyoto Free Translation Task (KFTT) test set~\citep{kftt}. Following \citet{sgnmt1} we measure time complexity in number of node expansions. Our depth-first search algorithm stops when a partial hypothesis score is worse than the current best complete hypothesis score (admissible pruning), but it is guaranteed to return the global best model score. Beam search yields a significant amount of search errors, even with a large beam of 20. Interestingly, a reduction in search errors does not benefit the BLEU score in this setting.

\begin{table}
\small
\centering

\begin{tabular}{|l|c|c|c|}
\hline
& \textbf{Average number of node} & \textbf{Sentences with} & \textbf{BLEU} \\
&  \textbf{expansions per sentence} & \textbf{search errors} & \textbf{score} \\
\hline
Exhaustive enumeration & 652.3K & 0\% & 21.7 \\
Depth-first search with admissible pruning & 3.0K & 0\% & 21.7 \\
Beam search (beam=20) & 250.5 & 20.3\% & 21.9 \\ 
Beam search (beam=4) & 64.8 & 41.9\% & 21.9 \\ 
Greedy decoding & 18.0 & 67.9\% & 22.1 \\
\hline
\end{tabular}
\caption{\label{tab:decoders} BPE-level SMT lattice rescoring with different search strategies. The BLEU score does not benefit from less search errors due to modeling errors.}
\end{table}

\section{Output Formats}
\label{sec:output-formats}

SGNMT supports five different output formats.

\begin{itemize}
\item \texttt{text}: Plain text file with first best translations.
\item \texttt{nbest}: $n$-best list of translation hypotheses.
\item \texttt{sfst}: Lattice generation in OpenFST~\citep{openfst} format with standard arcs.
\item \texttt{fst}: Lattices with sparse tuple arcs~\citep{sparsetuplearcs} which keep predictor scores separate.
\item \texttt{ngram}: MBR-style $n$-gram posteriors~\citep{mbr-smt,lmbr-tromble} as used by~\citet{mbr} for NMT.
\end{itemize}

\section{SGNMT for Research}
\label{sec:research}

\begin{table}
\small
\centering

\begin{tabular}{|l|c|c|c|}
\hline
& \textbf{Pure NMT} & \textbf{SMT lattice} & \textbf{MBR-based} \\
&  & \textbf{rescoring} & \textbf{NMT-SMT hybrid} \\
\hline
Theano: Blocks~\citep{blocks} & 18.4 & 18.9 & 19.0 \\
TensorFlow: seq2seq tutorial\footnotemark  & 17.5 & 19.3 & 19.2 \\
TensorFlow: NMT tutorial\footnotemark & 18.8 & 19.1 & 20.0 \\
TensorFlow: T2T Transformer~\citep{t2t} & 21.7 & 19.3 & 22.5 \\ \hline
\end{tabular}
\caption{\label{tab:kyoto} BLEU scores of SGNMT with different NMT back ends on the complete KFTT test set~\citep{kftt} computed with \texttt{multi-bleu.pl}. All neural systems are BPE-based~\citep{bpe} with vocabulary sizes of 30K. The SMT baseline achieves 18.1 BLEU.}
\end{table}

\addtocounter{footnote}{-1}
\footnotetext{\url{https://github.com/ehasler/tensorflow}}
\addtocounter{footnote}{+1}
\footnotetext{\url{https://github.com/tensorflow/nmt}, trained with Tensor2Tensor~\citep{t2t}}





SGNMT is designed for environments in which implementation time is far more valuable than computation time. This basic design decision is strongly reflected by the software architecture which accepts degradations in runtime in favor of extendibility and flexibility. We designed SGNMT that way because training models and coding usually take the most time in our day-to-day work. Decoding, however, usually takes a small fraction of that time. Therefore, reducing the implementation time has a much larger impact on the overall productivity of our research group than improvements in runtime, especially since decoding can be easily parallelized on multiple machines.

Another benefit of SGNMT's predictor framework is that it enables us to write code independently of any NMT package, and swap the NMT back end with more recent software if needed. For example, our previous research work on lattice rescoring~\citep{sgnmt1} and MBR-based NMT~\citep{mbr} used the NMT package Blocks~\citep{blocks} which is based on Theano~\citep{theano}. Since both Blocks and Theano have been  discontinued, we recently switched to a Tensor2Tensor~\citep{t2t} back end based on TensorFlow~\citep{tensorflow}. Without reimplementation, we could validate that MBR-based NMT holds up even under a much stronger NMT model, the Transformer model~\citep{transformer}. Tab.~\ref{tab:kyoto} compares the performance of lattice rescoring and MBR-based combination across four different NMT implementations using SGNMT.

\section{SGNMT for Teaching}
\label{sec:teaching}

SGNMT is being used for teaching at the University of Cambridge in course work and student research projects. In the 2015-16 academic year, two students on the Cambridge MPhil in Machine Learning, Speech and Language Technology used SGNMT for their dissertation 
projects. The first project involved using SGNMT with OpenFST~\citep{openfst} for applying subword models in SMT~\citep{mphil-jiameng}.   The second project developed automatic music composition by LSTMs where WFSAs were used to define the space of allowable chord progressions in `Bach' chorales~\citep{mphil-marcin}. The LSTM provides the `creativity' and the WFSA enforces constraints that the chorales must obey. This year, SGNMT provides the decoder for a student project about simultaneous neural machine translation.

SGNMT is also part of two practicals for MPhil students at Cambridge.\footnote{\url{http://ucam-smt.github.io/sgnmt/html/kyoto_nmt.html}} The first practical applies different kinds of language models to restore the correct casing in a lowercased sentence using FSTs. Since SGNMT has good support for the OpenFST library~\citep{openfst} and can both read and write FSTs, it is used to integrate neural models such as RNN LMs into the exercise. The second practical focuses on decoding strategies for NMT and explores the synergies of word- and subword-based models and the potential of combining SMT and NMT.

\section{SGNMT in the Industry}
\label{sec:industry}

SDL Research continuously balances the research and development of neural machine translation with a focus on bringing state-of-the-art MT products to the market\footnote{\url{http://www.sdl.com/software-and-services/translation-software/}} while pushing the boundaries of MT technology via innovation and quick experimental research.

In this context, it is highly desirable to use versatile tools that can be easily extended to support and combine new models, allowing for quick and painless experimentation. SDL Research chose SGNMT over all other existing tools for rapid prototyping and assessment of new research avenues. Among other Neural MT innovations, SDL Research used SGNMT to prototype and assess attention-based Neural MT~\citep{bahdanau}, Neural MT model shrinking~\citep{shrinking} and the recent Transformer model~\citep{transformer}. As described in Sec.~\ref{sec:research}, the Transformer model is trivially supported by the SGNMT decoder through its predictor framework, and is easy to combine with other predictors. It is worth noting that at the time of writing this paper, Transformer ensembles are not natively supported by the Tensor2Tensor decoder~\citep{t2t}.

Although SDL Research's decoder is homegrown, the SGNMT decoder is still a valuable reference tool for side-by-side comparison between state-of-the-art Neural MT research and the Neural MT product.

\small

\bibliographystyle{apalike}
\bibliography{mtsummit2015}

\begin{thebibliography}{}

\bibitem[Abadi et~al., 2016]{tensorflow}
Abadi, M., Agarwal, A., Barham, P., Brevdo, E., Chen, Z., Citro, C., Corrado,
  G.~S., Davis, A., Dean, J., Devin, M., et~al. (2016).
\newblock Tensorflow: {Large}-scale machine learning on heterogeneous
  distributed systems.
\newblock {\em arXiv preprint arXiv:1603.04467}.

\bibitem[Allauzen et~al., 2007]{openfst}
Allauzen, C., Riley, M., Schalkwyk, J., Skut, W., and Mohri, M. (2007).
\newblock {OpenFST: A general and efficient weighted finite-state transducer
  library}.
\newblock In {\em Implementation and Application of Automata}, pages 11--23.
  Springer.

\bibitem[Bahdanau et~al., 2015]{bahdanau}
Bahdanau, D., Cho, K., and Bengio, Y. (2015).
\newblock Neural machine translation by jointly learning to align and
  translate.
\newblock In {\em ICLR}, Toulon, France.

\bibitem[Bastien et~al., 2012]{theano}
Bastien, F., Lamblin, P., Pascanu, R., Bergstra, J., Goodfellow, I., Bergeron,
  A., Bouchard, N., Warde-Farley, D., and Bengio, Y. (2012).
\newblock Theano: new features and speed improvements.
\newblock In {\em NIPS}, South Lake Tahoe, Nevada, USA.

\bibitem[Bertoldi et~al., 2017]{mmt}
Bertoldi, N., Cattoni, R., Cettolo, M., Farajian, M., Federico, M., Caroselli,
  D., Mastrostefano, L., Rossi, A., Trombetti, M., Germann, U., et~al. (2017).
\newblock {MMT}: New open source {MT} for the translation industry.
\newblock In {\em Proceedings of The 20th Annual Conference of the European
  Association for Machine Translation (EAMT)}.

\bibitem[Chung et~al., 2016]{char-nmt}
Chung, J., Cho, K., and Bengio, Y. (2016).
\newblock A character-level decoder without explicit segmentation for neural
  machine translation.
\newblock In {\em Proceedings of the 54th Annual Meeting of the Association for
  Computational Linguistics (Volume 1: Long Papers)}, pages 1693--1703.
  Association for Computational Linguistics.

\bibitem[Gao, 2016]{mphil-jiameng}
Gao, J. (2016).
\newblock Variable length word encodings for neural translation models.
\newblock {MPhil} dissertation, University of Cambridge.

\bibitem[Gehring et~al., 2017]{convs2s}
Gehring, J., Auli, M., Grangier, D., Yarats, D., and Dauphin, Y.~N. (2017).
\newblock Convolutional sequence to sequence learning.
\newblock {\em ArXiv e-prints}.

\bibitem[Google, 2017]{t2t}
Google (2017).
\newblock {Tensor2Tensor: A} library for generalized sequence to sequence
  models.
\newblock \url{https://github.com/tensorflow/tensor2tensor}.
\newblock Accessed: 2017-12-12, version 1.3.1.

\bibitem[Hasler et~al., 2017]{eva-wordordering}
Hasler, E., Stahlberg, F., Tomalin, M., de~Gispert, A., and Byrne, B. (2017).
\newblock A comparison of neural models for word ordering.
\newblock In {\em Proceedings of the International Natural Language Generation
  Conference}, Santiago de Compostela, Spain.

\bibitem[Helcl and Libovick{\'{y}}, 2017]{neuralmonkey}
Helcl, J. and Libovick{\'{y}}, J. (2017).
\newblock {Neural Monkey: An} open-source tool for sequence learning.
\newblock {\em The Prague Bulletin of Mathematical Linguistics}, pages 5--17.

\bibitem[Hieber et~al., 2017]{sockeye}
Hieber, F., Domhan, T., Denkowski, M., Vilar, D., Sokolov, A., Clifton, A., and
  Post, M. (2017).
\newblock Sockeye: {A} toolkit for neural machine translation.
\newblock {\em ArXiv e-prints}.

\bibitem[Iglesias et~al., 2015]{sparsetuplearcs}
Iglesias, G., de~Gispert, A., and Byrne, B. (2015).
\newblock Transducer disambiguation with sparse topological features.
\newblock In {\em Proceedings of the 2015 Conference on Empirical Methods in
  Natural Language Processing}, pages 2275--2280. Association for Computational
  Linguistics.

\bibitem[Jean et~al., 2015]{rnnsearch-lv}
Jean, S., Cho, K., Memisevic, R., and Bengio, Y. (2015).
\newblock On using very large target vocabulary for neural machine translation.
\newblock In {\em Proceedings of the 53rd Annual Meeting of the Association for
  Computational Linguistics and the 7th International Joint Conference on
  Natural Language Processing (Volume 1: Long Papers)}, pages 1--10.
  Association for Computational Linguistics.

\bibitem[Junczys-Dowmunt et~al., 2016]{marian}
Junczys-Dowmunt, M., Dwojak, T., and Hoang, H. (2016).
\newblock Is neural machine translation ready for deployment? {A} case study on
  30 translation directions.
\newblock In {\em Proceedings of the 9th International Workshop on Spoken
  Language Translation (IWSLT)}, Seattle, WA.

\bibitem[Klein et~al., 2017]{opennmt}
Klein, G., Kim, Y., Deng, Y., Senellart, J., and Rush, A. (2017).
\newblock {OpenNMT: Open}-source toolkit for neural machine translation.
\newblock In {\em Proceedings of ACL 2017, System Demonstrations}, pages
  67--72. Association for Computational Linguistics.

\bibitem[Kumar and Byrne, 2004]{mbr-smt}
Kumar, S. and Byrne, W. (2004).
\newblock Minimum {Bayes}-risk decoding for statistical machine translation.
\newblock In {\em HLT-NAACL}, pages 169--176, Boston, MA, USA.

\bibitem[Luong et~al., 2015]{luong-attention}
Luong, T., Pham, H., and Manning, C.~D. (2015).
\newblock Effective approaches to attention-based neural machine translation.
\newblock In {\em Proceedings of the 2015 Conference on Empirical Methods in
  Natural Language Processing}, pages 1412--1421. Association for Computational
  Linguistics.

\bibitem[Neubig, 2011]{kftt}
Neubig, G. (2011).
\newblock The {Kyoto} free translation task.
\newblock http://www.phontron.com/kftt.

\bibitem[Neubig et~al., 2015]{nbest}
Neubig, G., Morishita, M., and Nakamura, S. (2015).
\newblock Neural reranking improves subjective quality of machine translation:
  {NAIST} at {WAT}2015.
\newblock In {\em WAT}, Kyoto, Japan.

\bibitem[Sennrich et~al., 2017]{nematus}
Sennrich, R., Firat, O., Cho, K., Birch, A., Haddow, B., Hitschler, J.,
  Junczys-Dowmunt, M., L{\"a}ubli, S., Miceli~Barone, A.~V., Mokry, J., and
  Nadejde, M. (2017).
\newblock Nematus: a toolkit for neural machine translation.
\newblock In {\em Proceedings of the Software Demonstrations of the 15th
  Conference of the European Chapter of the Association for Computational
  Linguistics}, pages 65--68. Association for Computational Linguistics.

\bibitem[Sennrich et~al., 2016]{bpe}
Sennrich, R., Haddow, B., and Birch, A. (2016).
\newblock Neural machine translation of rare words with subword units.
\newblock In {\em Proceedings of the 54th Annual Meeting of the Association for
  Computational Linguistics (Volume 1: Long Papers)}, pages 1715--1725.
  Association for Computational Linguistics.

\bibitem[Stahlberg and Byrne, 2017]{shrinking}
Stahlberg, F. and Byrne, B. (2017).
\newblock Unfolding and shrinking neural machine translation ensembles.
\newblock In {\em Proceedings of the 2017 Conference on Empirical Methods in
  Natural Language Processing}, pages 1946--1956. Association for Computational
  Linguistics.

\bibitem[Stahlberg et~al., 2017a]{mbr}
Stahlberg, F., de~Gispert, A., Hasler, E., and Byrne, B. (2017a).
\newblock Neural machine translation by minimising the {Bayes}-risk with
  respect to syntactic translation lattices.
\newblock In {\em Proceedings of the 15th Conference of the European Chapter of
  the Association for Computational Linguistics: Volume 2, Short Papers}, pages
  362--368. Association for Computational Linguistics.

\bibitem[Stahlberg et~al., 2017b]{sgnmt2}
Stahlberg, F., Hasler, E., Saunders, D., and Byrne, B. (2017b).
\newblock {SGNMT -- A} flexible {NMT} decoding platform for quick prototyping
  of new models and search strategies.
\newblock In {\em Proceedings of the 2017 Conference on Empirical Methods in
  Natural Language Processing: System Demonstrations}, pages 25--30.
  Association for Computational Linguistics.
\newblock Full documentation available at
  \url{http://ucam-smt.github.io/sgnmt/html/}.

\bibitem[Stahlberg et~al., 2016]{sgnmt1}
Stahlberg, F., Hasler, E., Waite, A., and Byrne, B. (2016).
\newblock Syntactically guided neural machine translation.
\newblock In {\em Proceedings of the 54th Annual Meeting of the Association for
  Computational Linguistics (Volume 2: Short Papers)}, pages 299--305.
  Association for Computational Linguistics.

\bibitem[Sutskever et~al., 2014]{sutskever}
Sutskever, I., Vinyals, O., and Le, Q.~V. (2014).
\newblock Sequence to sequence learning with neural networks.
\newblock In Ghahramani, Z., Welling, M., Cortes, C., Lawrence, N.~D., and
  Weinberger, K.~Q., editors, {\em Advances in Neural Information Processing
  Systems 27}, pages 3104--3112. Curran Associates, Inc.

\bibitem[Tomczak, 2016]{mphil-marcin}
Tomczak, M. (2016).
\newblock Bachbot.
\newblock {MPhil} dissertation, University of Cambridge.

\bibitem[Tromble et~al., 2008]{lmbr-tromble}
Tromble, R.~W., Kumar, S., Och, F., and Macherey, W. (2008).
\newblock Lattice minimum {Bayes}-risk decoding for statistical machine
  translation.
\newblock In {\em EMNLP}, pages 620--629, Honolulu, HI, USA.

\bibitem[van Merri{\"{e}}nboer et~al., 2015]{blocks}
van Merri{\"{e}}nboer, B., Bahdanau, D., Dumoulin, V., Serdyuk, D.,
  Warde{-}Farley, D., Chorowski, J., and Bengio, Y. (2015).
\newblock Blocks and fuel: Frameworks for deep learning.
\newblock {\em CoRR}.

\bibitem[Vaswani et~al., 2017]{transformer}
Vaswani, A., Shazeer, N., Parmar, N., Uszkoreit, J., Jones, L., Gomez, A.~N.,
  Kaiser, L.~u., and Polosukhin, I. (2017).
\newblock Attention is all you need.
\newblock In {\em Advances in Neural Information Processing Systems 30}, pages
  6000--6010. Curran Associates, Inc.

\bibitem[Vaswani et~al., 2013]{nplm}
Vaswani, A., Zhao, Y., Fossum, V., and Chiang, D. (2013).
\newblock Decoding with large-scale neural language models improves
  translation.
\newblock In {\em Proceedings of the 2013 Conference on Empirical Methods in
  Natural Language Processing}, pages 1387--1392, Seattle, Washington, USA.
  Association for Computational Linguistics.

\bibitem[Williams et~al., 2014]{uedin-wmt14}
Williams, P., Sennrich, R., Nadejde, M., Huck, M., Hasler, E., and Koehn, P.
  (2014).
\newblock Edinburgh’s syntax-based systems at {WMT} 2014.
\newblock In {\em Proceedings of the Ninth Workshop on Statistical Machine
  Translation}, pages 207--214, Baltimore, Maryland, USA. Association for
  Computational Linguistics.

\bibitem[Wu et~al., 2016]{gnmt}
Wu, Y., Schuster, M., Chen, Z., Le, Q.~V., Norouzi, M., Macherey, W., Krikun,
  M., Cao, Y., Gao, Q., Macherey, K., et~al. (2016).
\newblock Google's neural machine translation system: {Bridging} the gap
  between human and machine translation.
\newblock {\em arXiv preprint arXiv:1609.08144}.

\end{thebibliography}

\end{document}